\documentclass{article}

\usepackage{microtype}
\usepackage{graphicx}
\usepackage{booktabs} 

\usepackage{hyperref}
\usepackage{times}
\usepackage{latexsym}

\usepackage{microtype}
\usepackage{booktabs} 
\usepackage{breakurl}

\usepackage{amsmath}
\usepackage{graphicx}
\usepackage[export]{adjustbox}
\usepackage{caption}
\usepackage{subcaption}
\usepackage{algorithm}
\usepackage{algorithmic}
\usepackage{multirow}
\usepackage{enumitem}
\usepackage{amsfonts}
\usepackage{float}
\usepackage[flushleft]{threeparttable}

\usepackage[accepted]{mlsys2020}

\mlsystitlerunning{Doping: A technique for Extreme Compression of LSTM Models using Structured Matrices}

\begin{document}

\twocolumn[
\mlsystitle{Doping: A technique for Extreme Compression of LSTM Models using Sparse Structured Additive Matrices}




\begin{mlsysauthorlist}
\mlsysauthor{Urmish Thakker}{sn}
\mlsysauthor{Paul N. Whatmough}{arm}
\mlsysauthor{Zhigang Liu}{arm}
\mlsysauthor{Matthew Mattina}{arm}
\mlsysauthor{Jesse Beu}{arm}
\end{mlsysauthorlist}

\mlsysaffiliation{arm}{Arm ML Research}
\mlsysaffiliation{sn}{SambaNova Systems}

\mlsyskeywords{Deep Learning, Structured Matrix, Model Compression}

\mlsyscorrespondingauthor{Urmish Thakker}{uthakker@cs.wisc.edu}

\vskip 0.3in

\begin{abstract}

Structured matrices, such as those derived from Kronecker products (KP), are effective at compressing neural networks, but can lead to unacceptable accuracy loss when applied to large models.
In this paper, we propose the notion of \textit{doping} - addition of an extremely sparse matrix to a structured matrix. Doping facilitates additional degrees of freedom for a small number of parameters, allowing them to independently diverge from the fixed structure. 
To train LSTMs with doped structured matrices, 
we introduce the additional parameter matrix while slowly annealing its sparsity level.
However, we find that performance degrades as we slowly sparsify the doping matrix, due to co-matrix adaptation (CMA) between the structured and the sparse matrices.
We address this over dependence on the sparse matrix using a co-matrix dropout regularization (CMR) scheme. We provide empirical evidence to show that doping, CMA and CMR are concepts generally applicable to multiple structured matrices (Kronecker Product, LMF, Hybrid Matrix Decomposition). Additionally, results with doped kronecker product matrices demonstrate state-of-the-art accuracy at large compression factors ($10-25\times$) across 4 natural language processing applications with minor loss in accuracy. Doped KP compression technique outperforms previous state-of-the art compression results by achieving $1.3-2.4\times$ higher compression factor at a similar accuracy, while also beating  strong alternatives like pruning and low-rank methods by a large margin (8\% or more). Additionally, we show that doped KP can be deployed on commodity hardware using the current software stack and achieve $2.5-5.5\times$ inference run-time speed-up over baseline.

\end{abstract}

]



\printAffiliationsAndNotice{}  

\section{Introduction}

Language models (LMs) based on neural networks have been extremely effective in enabling a myriad of natural language processing (NLP) applications in recent years.
However, many of these NLP applications are increasingly being deployed on consumer mobile devices and smart home appliances, where the very large memory footprint of large LMs is a severe limitation.
To help bridge this gap in model size, model optimization techniques have been demonstrated to reduce the memory footprint of the weight matrices~\cite{igor1,DennisAVSSJ19,KusupatiSBKJV18}.
However, results published to date still result in very significant off-chip DRAM bandwidth, which increases the power consumption of mobile devices~\cite{li-dac19}.
This is especially significant for NLP applications that are running for long periods of time.
For example, to efficiently deploy a 25 MB LM on a device with 1 MB L2 cache, requires 25$\times$ compression or a 96\% reduction in the number of parameters. 
Such a high pruning ratio leads to significant accuracy degradtion~(\cite{stateSparse,suyog}), which is not acceptable from an application point of view.
Therefore, to enable efficient inference on resource constrained devices~\cite{zhu-mlsys19}, there is a sustained need for improved compression techniques that can achieve high compression factors without compromising accuracy.

\begin{figure*}[t]
\centering

\vspace{8pt}

\begin{subfigure}[b]{\columnwidth}
\centering
\includegraphics[width=0.8\columnwidth]{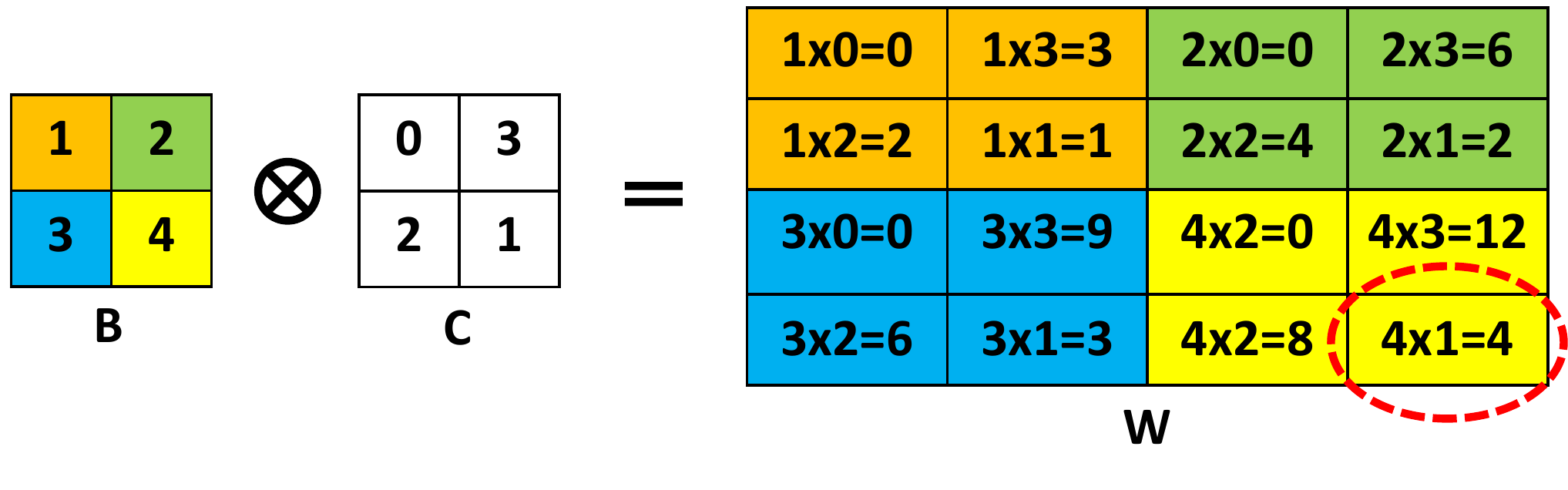}
\caption{Conventional Kronecker product matrix.}
\label{dkp-fig:kp}
\end{subfigure}
\begin{subfigure}[b]{\columnwidth}
\centering
\includegraphics[width=1.0\columnwidth]{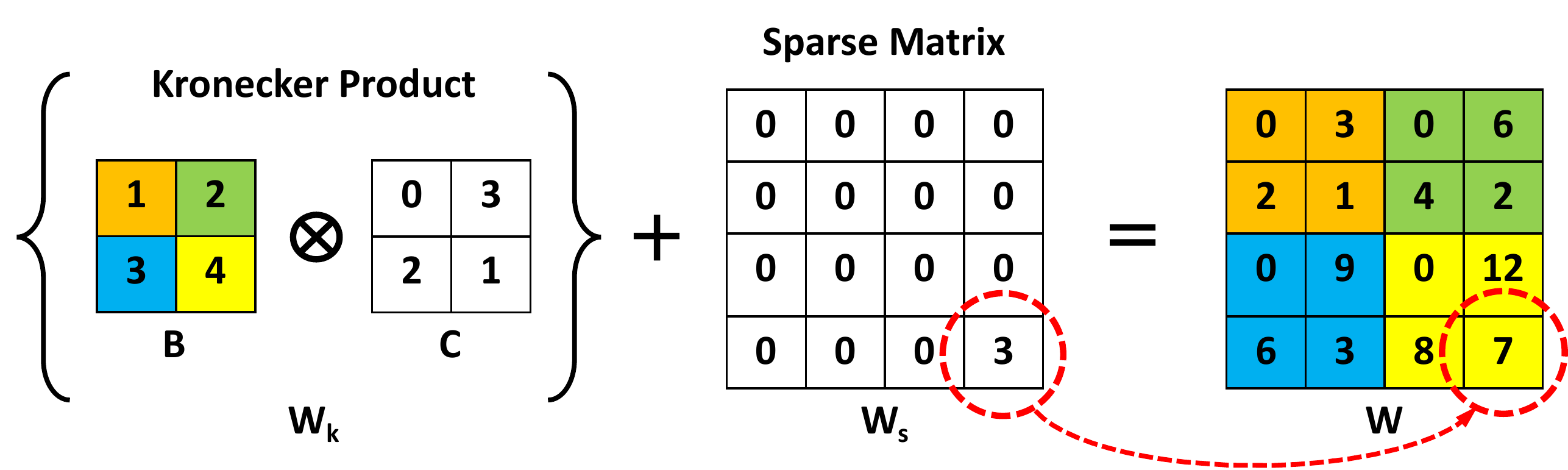}
\caption{Doped Kronecker product matrix.}
\label{dkp-fig:dkp}
\end{subfigure}

\caption{
In the conventional Kronecker product (a), individual elements in the expansion $W$ cannot diverge independently without impacting other elements in the block.
This restricts freedom in the parameter space and limits accuracy.
By \textit{doping} with a sparse matrix $M_{sp}$ (b), we relax this constraint with an additional degree of freedom in the parameter space, with minimal overhead.
}
\label{dkp-fig}

\vspace{8pt}

\end{figure*}

Model compression using structured matrices has been previously demonstrated on various image and audio tasks, such as object detection, human activity recognition (HAR), orthogonal projections and key-word spotting (KWS) \cite{stanStruct,circular2,circ3,permDNN,Thakker2019PushingTL}. 
For example, Kronecker Products (KP) were used to compress HAR and KWS models by 15-38$\times$ compression factors \cite{urmkron2} while achieving better accuracy than other traditional compression techniques such as pruning \cite{suyog} and low-rank matrix factorization (LMF).
However, applying KP to more complex tasks, such as a large LM, results in a 26\% loss in perplexity, while applying a low rank structure using tensor train decomposition leads to a greater than 50\% decrease in perplexity score \cite{2019_c3}. 
Figure~\ref{dkp-fig:kp} shows that the root issue with conventional KP is that during the optimization of the constituent matrices $B$ and $C$, all elements of $W$ inside each block of 4 elements (colored in the figure) are related.
This limitation in the parameter space leads to perplexity loss on larger problems. Specifically, the issue gets exaggerated in bigger matrices as structured decomposition of bigger matrices for large compression factors creates more number of such relations.
Other structured matrices~\cite{stanStruct} face a similar challenge. 

In this paper, we relax this limitation in the parameter space by \textit{doping} the structured matrix $M_{k}$ with an extremely sparse additive matrix $M_{s}$ (Figure~\ref{dkp-fig:dkp}).
This approach was inspired by robust PCA techniques, and allows an additional degree of freedom to recover accuracy at very low inference-time cost.
The contributions of this paper are further summarized below:
\setlist{nolistsep}
\begin{itemize}[noitemsep]
    \item We propose \textit{sparse matrix doping}\footnote{The term ``doping'' is an analogy to intentionally introducing impurities into an intrinsic semiconductor in material science.}, which addresses the accuracy limitations of structured matrix compression using an additional sparse matrix with negligible inference cost.
    \item A training recipe to learn the structured and the sparse matrix, describing regularization techniques to reduce the co-matrix adaptation (CMA) encountered as we anneal sparsity, 
    using co-matrix regularization (CMR).
    \item Show that doping, CMA and CMR are applicable to a a wide variety of structured matrices.
    \item Present state-of-the-art results for compressing language models and translation applications using doped KP, which are 1.5$\times$ - 2.4$\times$ smaller than previous work at comparable accuracy values.
\end{itemize}

The remainder of the paper is organized as follows.
Section~\ref{sec:related} briefly surveys related work.
Section \ref{sec:dkp}, introduces sparse matrix doping by applying the technique to Kronecker product matrix compression, which is the main focus of the paper.
We discuss the challenges in training doped structured matrices (Section \ref{sec-cma}) and propose a training recipe that includes regularization to prevent co-matrix adaptation which otherwise leads to accuracy loss (Section \ref{sec-cmr}).
Using this training recipe, we discuss how doped structured matrix compression techniques can be applied to a variety of structured matrices in Section~\ref{sec:dopedlmf} and discuss why doped KP leads to higher accuracy than other structured matrices evaluated in this paper.
We demonstrate state-of-the art models using doped KP compression on three language modeling tasks and one translation task in Sections \ref{res:mediumlm},\ref{res:mediumrhn} and \ref{res:gnmt}.
Doping introduces negligible run-time overhead, which we confirm by benchmarking the inference performance of all models on a Raspberry Pi 4 (Section~\ref{res:inf}).

\section{Related Work}
\label{sec:related}

\textbf{Random Pruning} \cite{han2015deep_compression,suyog,sanh2020movement,fedorov2020tinylstms,whatmough-jssc18,dbb,fixy2019sysml} of neural networks seems to be a very successful compression technique across many different tasks. However, so-called random pruning is hard to exploit on real hardware due to the lack of regularity in the resulting matrix multiplications unless the sparsity level is large. 

\textbf{Structured Matrices} have shown significant potential for compression of neural networks~\cite{circular2,structuredmatrix,circnn,circular1,kron3,stanStruct,urmkron2,dkp}. Block circular compression is an extension of structured matrix based compression technique, converting every block in a matrix into a structured matrix. Building on top of this, our work proposes a method to increase the compression achieved using structured matrices at baseline accuracy by introducing sparse matrix doping on top of structured matrices.

\textbf{Tensor Decomposition} including Tucker decomposition, Kronecker etc. These methods have also been employed to achieve significant reduction in parameters \cite{tjandra2017compressing,GopeMLSys2019,TernaryMobileNetstinyMLSummit2020}. Low-rank Matrix Factorization (LMF) \cite{DBLP:journals/corr/KuchaievG17,lmf-bad,lmf-good1,urmtha01-hmd,thakker2020rank} can also be categorized under this topic. LMF aware NN are a special case of structured matrix as it also imposes a certain constraint on the expressibility of the matrix. We will show in this paper that doped KP can lead to better accuracy than LMF.

\textbf{Quantization} is another popular technique for compression \cite{Quant-hubara,Quant-bengio,sanh2019distilbert,bin-lstm,Gope2019TernaryMV,banbury2021micronets}. Networks compressed using DKP can be further compressed using quantization.

\textbf{Dynamic techniques} are used to improve inference run-time of RNNs by skipping certain RNN state updates \cite{skiprnn,skimrnn,lstmjump,urmSkipRNN}. These techniques are based on the assumption that not all inputs to an RNN are needed for final classification task. Thus we can learn a small and fast predictor that can learn to skip certain inputs and its associated computation. Doped KP technique is orthogonal to this technique and networks compressed using doped KP can be further optimized using this technique. These dynamic techniques could be extended to networks beyond LSTMs also \cite{ravi,annie}.

\textbf{Efficient Network Architectures} for LSTMs, such as SRU \cite{sru}, QRNN \cite{qrnn} and PRU \cite{pru} have also led to networks with faster inference run-time performance through increased parameter efficiency. Doped KP can be used to compress the weight matrices in all of these architectures to further optimize the inference time performance.

\textbf{Word Embedding Compression} is another way to reduce the parameter footprint of embedding matrices in NLP \cite{embed1,embed2}. In this paper, we show that DKP can compress networks with compressed word embedding layers also. 

\textbf{Combining features:} The doped KP compression technique replaces a weight matrix with the sum of a sparse additive matrix and a structured matrix. 
This leads to output features that are a combination of those generated from a structured matrix and another from a sparse matrix. \citet{KusupatiSBKJV18} and \citet{resnet} also combine features from two different paths. 
However, their work deviates from ours for multiple reasons. 
Firstly, their aim is to combine features to tackle the vanishing gradients issue. 
Secondly, they use a residual connection for the additive feature, and as a result, they do not see the CMA issue discussed in our work. 
Finally, their technique is not well suited for the purpose of enabling additional degrees of freedom in features constrained by a structured matrix. 
Specifically, the sparse matrix selects specific elements in a feature vector that need additional degrees of freedom which their work cannot do.

In this work, doped KP combines very high-sparsity random pruning with structured matrix techniques. Our results are compared with pruning, structured matrix and tensor decomposition techniques. Quantization can subsequently be used to further compress the models we present.

\section{Doped Kronecker Product}
\label{sec:dkp}

In this paper we will focus on applying the doping technique to Kronecker product structured matrices.
However, doping can be generalized to other structured matrices and we will also give some results for other structures in Section~\ref{sec:results}.

Let lowercase and uppercase symbols denote vectors and matrices, respectively.
In doped KP, a parameter matrix $W$ is the sum of a KP matrix $W_{k}$ and a sparse matrix $W_{s}$ (Figure~\ref{dkp-fig:dkp}),
\begin{align}
    W = W_{k} + W_{s},
    \label{eq:dkp} \\
    W_{k} = B\otimes C,
\end{align}
where, $\otimes$ represents the Kronecker operator \cite{kpweb}
For $B \in \mathbb{R}^{ M1\times N1}$, $C \in \mathbb{R}^{ M2\times N2}$,  $W \in \mathbb{R}^{M \times N}$, $M = M1\times M2$ and $N = N1\times N2$ the compression factor (CF) can be calculated using the formula:
\begin{gather}
CF = (M1*N1 + M2*N2 + \left\lVert W_{s} \right\lVert _0 )/ (M*N).
\end{gather}

\subsection{Sizing doped KP Matrices}
\label{sec:kpparam}
Identifying the dimensions of the $B$ and $C$ matrices is non-trivial. A matrix can be expressed as a KP of multiple smaller matrices of varying sizes. For example, if $\boldsymbol{W_{k}}$ is of size 100$\times$100, B and C can be of size $2\times50$ and $50\times 2$ each or $10\times10$ and $10\times 10$ each. Both solutions lead to a compression factor of $50\times$. Recent work~\cite{urmkron2} described a straightforward methodology to achieve maximum compression of $W_{k}$ using KP, while achieving maximum rank. Empirical evidence suggests that the large rank value corresponds to larger accuracy gains. Therefore, we adopt the methodology in \cite{urmkron2} to decompose $W_{k}$ into $B$ and $C$. We run ablation studies to further confirm that this choice indeed holds true. The results of these ablation studies can be found in Appendix~\ref{sec:kpchoice}.

Once the size of $W_{k}$ is fixed, $\left\lVert W_{s} \right\lVert _0$
can be set during hyperparameter optimization,
to maximize the compression factor without impinging on performance. As an example, if $\boldsymbol{W}$ is of size 100$\times$100, $\boldsymbol{B}$ and $\boldsymbol{C}$ are of size 10$\times$10, then 95\% sparsity in $\boldsymbol{W}_{s}$ results in a compression factor of 14$\times$; the same scenario with 90\% sparsity in $\boldsymbol{W_{s}}$ results in an 8.4$\times$ compression factor.

\subsection{Training doped KP Networks}

The doped KP networks are trained from scratch, i.e. they do not start from a pre-trained network. Doped KP replaces all the parameter matrices in the neural network with the sum of $W_{k}$ and $W_{s}$ matrices. We initially set $W_s$ with a dense random initialization. Then, as training progresses, we apply magnitude weight pruning to $W_s$, annealing the sparsity towards the target level. We use the methodology described in \cite{suyog} to achieve this transition, which aims to allow the training optimization algorithm to retain the non-zero elements in $W_{s}$, which have the greatest impact on cross entropy.

\begin{figure}[t]
\centering
\includegraphics[width=\columnwidth]{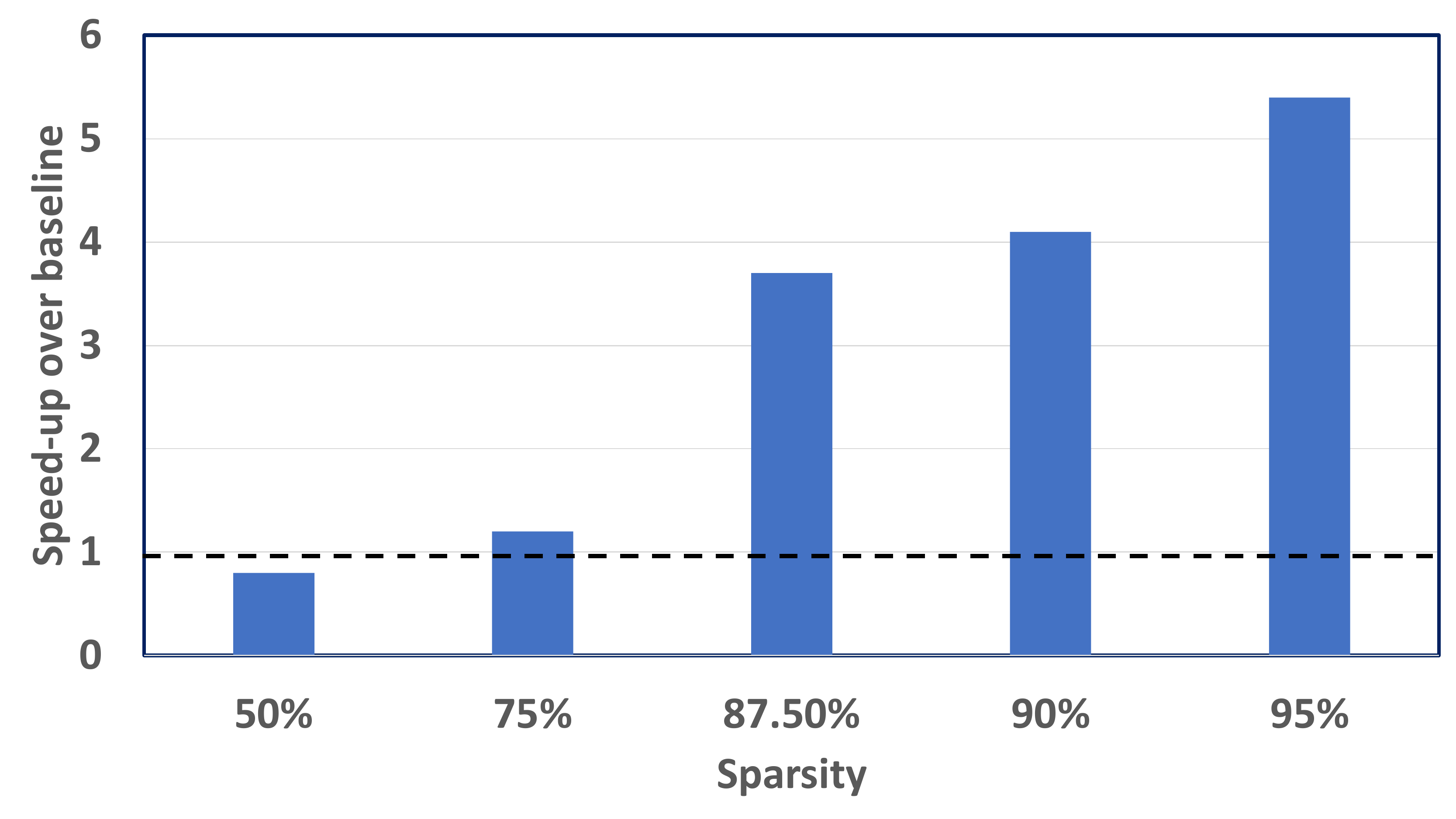}
\vskip -0.1in
\caption{
Speed-up of matrix-vector product kernels as a function of matrix sparsity, with a dense vector.
The matrix is of dimension $256\times 256$. 
Measured on a single Arm Cortex A-72 CPU of the Raspberry Pi 4 board using the Eigenc C++ library.
}
\label{fig:sparsemv}
\vskip -0.1in
\end{figure}

\subsection{Inference on doped KP Networks}

For inference on an edge device, NLP applications generally use a batch size of one and thus execute a matrix-vector product kernel \cite{urmtha01RNN}. The matrix vector product for doped KP will lead to the execution of the following computation:
\begin{align}
    y &= W*x \\
    y &= (W_{k} + W_{s})*x,\; where
    \label{eq:dkpmv} \\
    W_{k} &= B\otimes C,
\end{align}
where,$W\in R^{M\times N}$, $x\in R^{N\times 1}$, $y\in R^{M\times 1}$, $B \in R^{M1\times N1}$, $C \in R^{M2\times N2}$, $M1\times M2 = M$ and $N1\times N2 = N$.

The above inference leads to multiply accumulate (MAC) and actual runtime reductions as both $W_{k}*x$ and $W_{s}*x$ can be computed cheaply.
\paragraph{Inference Cost of ($W_{k}*x)$:} \cite{urmkron2} show that to execute $W_{k}*x$ on an embedded hardware, we do not need to expand the Kronecker matrix to the larger matrix. We can achieve significant speedup by using the following set of equations \cite{kpweb} -
\begin{align}
    y_{k} &= (B\otimes C)*x \\
    y_{k} &= vec(Y_{k}) \; where,\\
    Y_{k} &= C\times B\times X^{T} \; and\; X\;=\;matrix(x)
\end{align}
where $X \in R^{N2\times N1}$, $Y_{k}\in R^{M2\times M1}$, $vec()$ converts a matrix into a vector and $matrix()$ converts a vector into matrix. Thus, the matrix vector product, when the matrix is expressed as KP of two smaller matrices, gets converted into two small GEMM kernel calls. 

\paragraph{Inference Cost of ($W_{s}*x)$:} We emphasize that the additional latency cost of doping by adding $W_{s}*x$ is negligible at inference time as long as $W_s$ is sufficiently sparse. Figure~\ref{fig:sparsemv} shows the results of running matrix-vector product calculation on a Raspberry Pi 4 development board \cite{rasp}, for various sparsity values of the matrix. The matrix is of dimension $256\times256$ and the kernels use Eigen C++ library \cite{eigen}. 
Clearly, the matrix sparsity has to be rather high to achieve a speedup compared to the dense baseline.
However, with doping, the additional sparse matrix can be very high and therefore the additional cost is low.
For example, in this paper we target $W_{s}$ sparsity values of $10\times$ or more, as a result the matrix-vector product kernel can be executed at a speed-up of $4.1\times$ or more when compared against the baseline implementation.


\begin{figure}[tb]
\vspace{-5pt}
\centering
\includegraphics[width=\columnwidth]{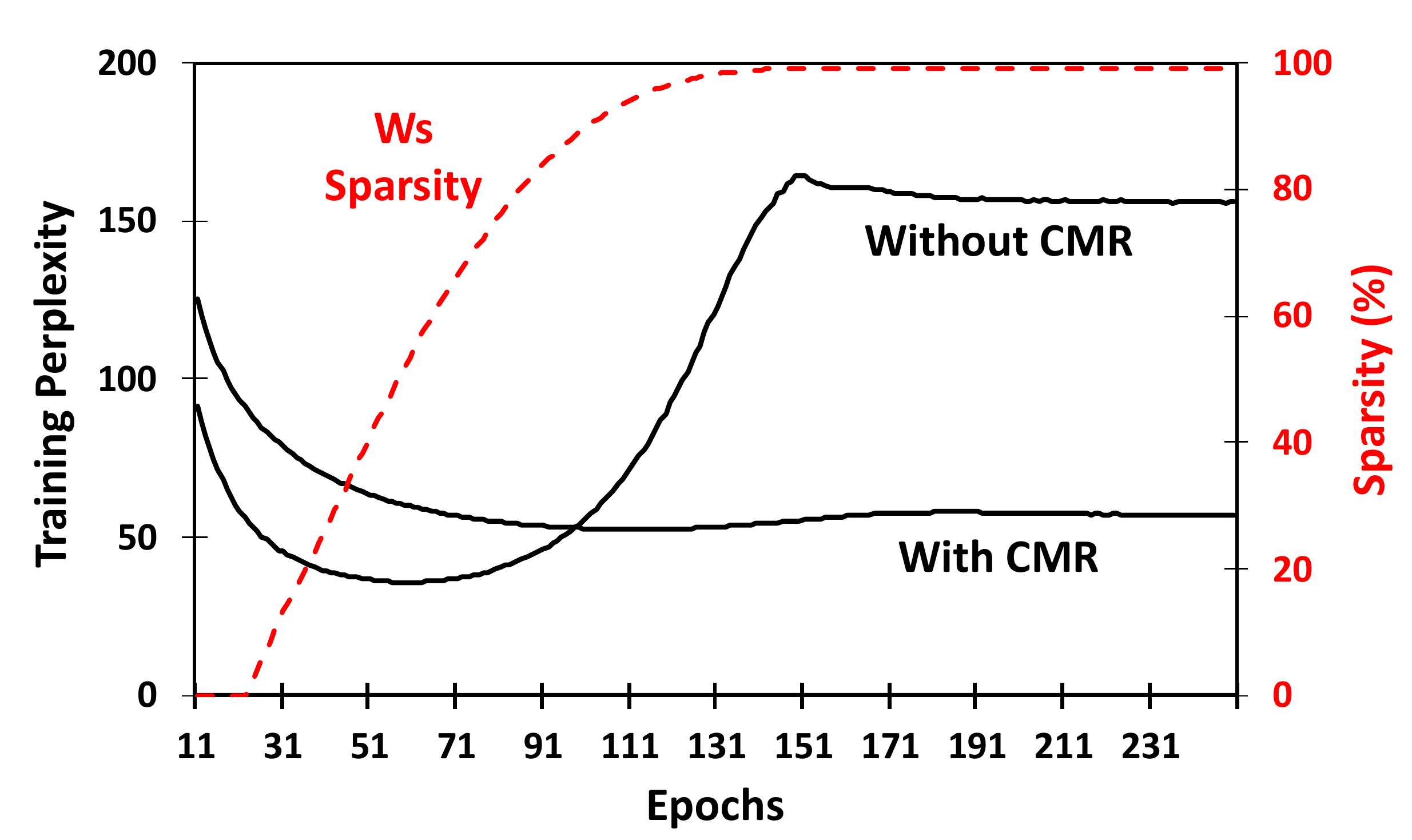}
\vspace{-21pt}
\caption{
Training a medium LM at 100$\times$ compression.
As the sparsity of $W_s$ is increased, training perplexity degrades due to over dependence on non-zero elements of $W_{s}$ established early on.
Using CMR we are able to prevent perplexity collapse as we anneal sparsity.
}
\label{cma-fig}
\end{figure}

\section{Co-Matrix Adaptation (CMA)}
\label{sec-cma}

Initial attempts to train doped KP LSTMs were hampered by accuracy collapse as $W_s$ sparsity was increased.
For example, we compressed the medium LM from~\cite{ptblm} using KP \cite{urmkron2} and doped KP (Rows 2 and 3 in Table \ref{tab:cma-test-main}). 
However, the resulting perplexity score degrades by 44.7\% when trained using doped KP, even with slightly more parameters.
To clarify the source of this accuracy loss, Figure~\ref{cma-fig} shows the training perplexity and sparsity for the medium LM at an aggressive 100$\times$ CF to clearly illustrate the issue. 
As the sparsity of $W_{s}$ increases, the perplexity soon collapses,
indicating that the model develops an over reliance on $W_{s}$ while it is dense, during the initial training phase.
Subsequently, as we anneal the sparsity of $W_{s}$, the perplexity collapses.
We refer to this phenomena as co-matrix adaptation (CMA).

\subsection{Understanding CMA}
An input feature vector $x$ in a dense layer is multiplied with the doped KP weight matrix (eq~\ref{eq:dkp}) to give the output $y$,
\begin{gather}
    y = W_{k} x + W_{s} x.
\end{gather}
Thus, the input flows through both $W_{k}$ and $W_{s}$ and the output of the matrix-vector product is combined to create the final output. The above equation can be viewed as,
\begin{equation}
    y^{j} = (w_{k}^{j})^T x + (w_{s}^{j})^T x,
    \label{eq:cmaoutput}
\end{equation}
where the superscript refers to the $j^{th}$ row of the matrix.
Each element of the output feature vector is the sum of neurons coming from the $W_{k}$ matrix and the $W_{s}$ matrix. CMA arises early on due to the co-adaptation of the incoming neurons through both $W_{k}$ and $W_{s}$, essentially balancing the significance of the each.
However, this is problematic as we start to progressively prune $W_{s}$ as training progresses. 

This co-adaptation is also obvious when we focus on the number of back-propagation updates during the initial phase of training. The rate of back-prop updates is not even between $W_k$ and $W_s$, which introduces further undesirable emphasis on the dense $W_s$. In the example medium LM, $B$ is 52$\times$65, and $C$ is 50$\times$20, and therefore $W_k$ has a total of $\sim$4.4K parameters.
While $W_{s}$ is 2600$\times$1300, which is $\sim$3.4M parameters in the initial dense form.
Thus, during the initial training phase, $W_{s}$ receives 700$\times$ more weight updates than $W_{k}$ leading to the over-reliance. 

\begin{table}[tb]
\centering
\begin{tabular}{lccc}
\toprule
\begin{tabular}[c]{@{}l@{}}Compression\\Factor\end{tabular} & \begin{tabular}[c]{@{}l@{}}Training\\Method\end{tabular} & \begin{tabular}[c]{@{}l@{}}Sparsity\\of $M_{sp}$\end{tabular} & \begin{tabular}[c]{@{}l@{}}Test\\Ppl.\end{tabular} \\ 
\midrule
1x & - & - & 82.1 \\ 
\midrule
$338\times$ (KP) & - & 0 & 104.1 \\ 
\midrule
\multirow{7}{*}{$100\times$ (doped KP)} & Eq 1 & 99\% & 150.7 \\ 
 & Eq \ref{eq:dkpcma120}+BCD & 99\% & 100.5 \\ 
 & Eq \ref{eq:dkpcma3456}+BCD & 99\% & 100.9 \\ 
 & CMR & 99\% & 95.4 \\ 
 \bottomrule
\end{tabular}
\caption{Medium LM 100$\times$ compression results using KP and DKP. Four different DKP networks are evaluated: one trained without CMR (eq~\ref{eq:dkp}), others trained using eq~\ref{eq:dkpcma120}, eq~\ref{eq:dkpcma120} with block coordinate descent (BCD) and CMR (eq~\ref{eq:cmr}). While other techniques show promising results, CMR achieves drastically improved perplexity showing that it is a more effective training technique for doped KP networks.
}
\label{tab:cma-test-main}
\vspace{-1em}
\end{table}
\section{Co-Matrix Regularization (CMR)}
\label{sec-cmr}

In order to prevent undesirable reliance on non-zero elements of $W_{s}$ that will later be pruned away, we introduce a random row dropout process which we refer to as co-matrix regularization (CMR). 
Thus, eq~\ref{eq:cmaoutput} becomes
\begin{equation}
    y^{j} = ((\textit{w}_{k}^{j})^T \textit{x})\circ b_1 + ((\textit{w}_{s}^{j})^T \textit{x} )\circ b_2,
    \label{eq:cmr}
\end{equation}

where $b_{1}$ and $b_{2}$ are CMR dropout values drawn from Bernoulli distribution with probability $p$, and $\circ$ is element-wise multiplication.
CMR creates 4 different scenarios for the output neuron:

\begin{align}
    &\; y^{j} = (\textit{w}_{k}^{j})^T \textit{x} + (\textit{w}_{s}^{j})^T \textit{x} \\
    &\; y^{j} = (\textit{w}_{k}^{j})^T \textit{x} \\
    &\; y^{j} = (\textit{w}_{s}^{j})^T \textit{x} \\
    &\; y^{j} = 0 \;(regular\;dropout)
\end{align}

By ensuring that the output neuron is occasionally produced without the presence of one of the incoming neurons, CMR can reduce the co-dependence between the matrices.
Figure~\ref{cma-fig} demonstrates that CMR helps manage CMA and prevent perplexity collapse,
Table \ref{tab:cma-test-main} summarizes final test accuracy, which 
is improved by 36.7\% with CMR (equation~\ref{eq:cmr}). 

\subsection{Alternatives to CMR}
\label{sec:altcmr}
We also explore other ways to manage CMA that relied on adding constraints to the $W_{k}$ and $W_{s}$ matrices in varying forms:  
\begin{subequations}
\label{eq:dkpcma}
\begin{align}
        W&=B \otimes C + \beta \times W_{s},  min \left\lVert \beta \right\rVert      
        \label{eq:dkpcma120}
\end{align}
\begin{align}
        \begin{split}
            W =\alpha \times (B \otimes C) + \beta \times W_{s}, \\
            min (\left\lVert \beta \right\rVert + \left\lVert 1/\alpha \right\rVert)        \label{eq:dkpcma3456}
        \end{split}        
\end{align}
\end{subequations}

Each of these constrains in equation \ref{eq:dkpcma120} - \ref{eq:dkpcma3456} can be further enhanced by training using Block Coordinate Descent (BCD). In BCD we alternate between, only training $W_{k}$, blocking gradient flow to $W_{s}$, or train $W_{k}$, blocking gradient flow to $W_{s}$. 

\begin{figure}[t]
\vskip 0.1in
\centering
\includegraphics[width=\columnwidth]{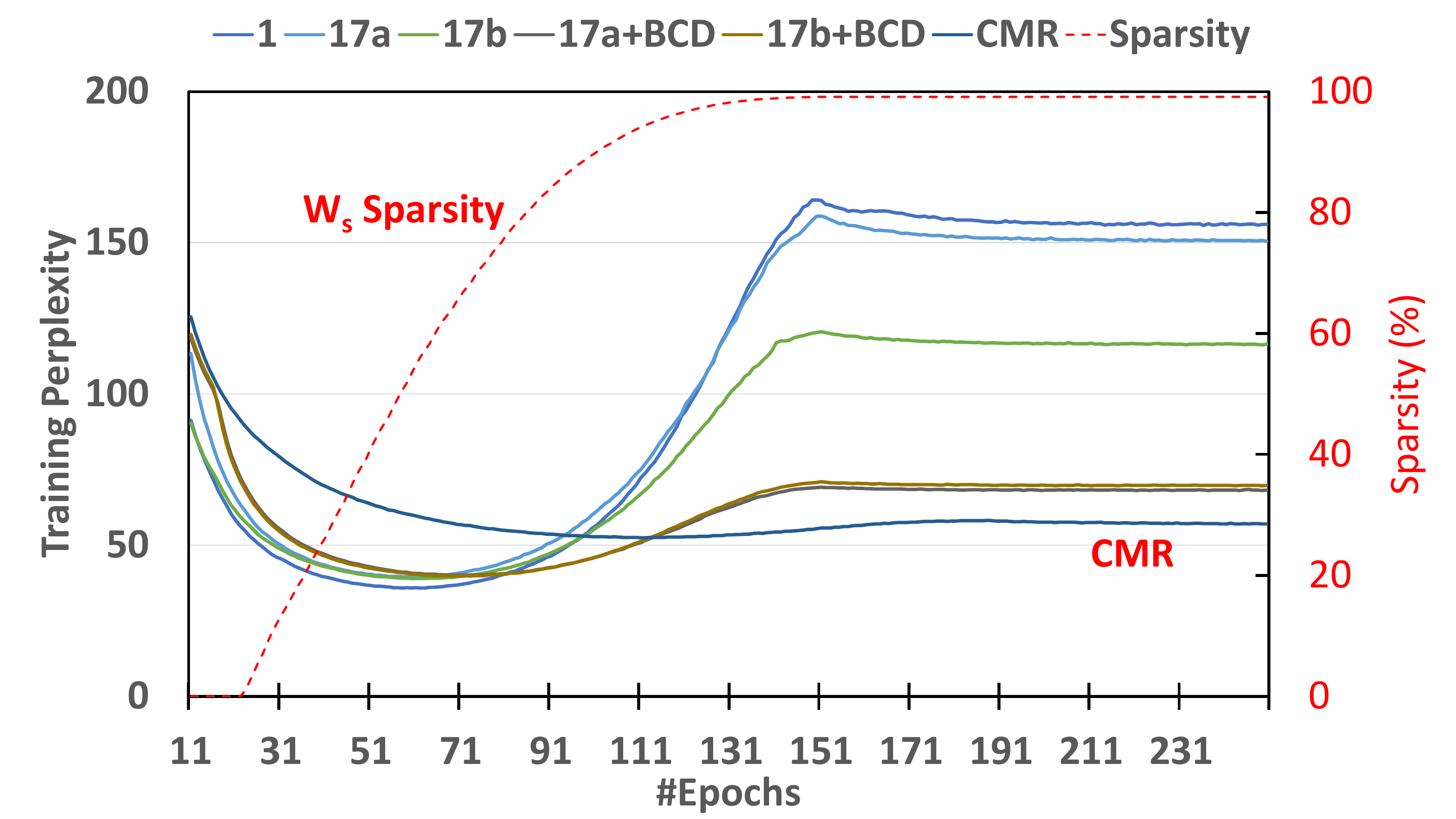}
\vskip -0.1in
\caption{(Best viewed in color) Using the various alternative techniques described in equation \ref{eq:dkpcma120} - \ref{eq:dkpcma3456} with and without block coordinate descent (BCD), we see that the reliance on $W_{s}$ has reduced. The increase in training perplexity that was visible in Figure \ref{cma-fig} has been managed considerably. However, none of the alternative techniques match the training accuracy of CMR.}
\label{overcomecma-fig}
\vskip -0.1in
\end{figure}

The training curves and test results corresponding to these methods of overcoming CMA are shown in Figure \ref{overcomecma-fig} and Table \ref{tab:cma-test-main}. While these methods help in overcoming CMA, CMR is far more effective than these techniques.  


\subsection{CMR for Generalized Doped Structured Matrices}

CMR is a phenomenon that we observe when we combine $W_{s}$ with other structured matrices also. We run experiments where we combine a low-rank matrix factorized matrix \cite{DBLP:journals/corr/KuchaievG17} with a sparse matrix, i.e., the parameter matrix is replaced as:

\begin{align}
    W = W_{lmf} + W_{s},
    \label{eq:dlmf} \\
    W_{lmf} = B\times C,
\end{align}

where, $B \in R^{M\times d}$, $C \in R^{d\times N}$ and $W_{lmf} \in R^{M\times N}$ and $d < min(M,N)$. We call this method doped LMF. Similarly, we can create doped HMD \cite{urmtha01-hmd} compression method. We show that doping, CMR and CMA are applicable and useful for achieving high-accuracy for doped LMF (HMD) compression method also. Section \ref{sec:dopedlmf} discusses results of compression using doped LMF (HMD) and how they fare against doped KP compression method.

\subsection{Training using CMR Dropout}
\label{sec:ablcmr}
CMR dropout is needed to avoid CMA. However as the $W_{s}$ gets pruned over-time, the need for CMR decreases. 
In order to verify this, we experiment with 3 different CMR schedules:
\begin{itemize}
    \item \textbf{constant}: The CMR dropout value remains constant throughout the training run
    \item \textbf{linDec}: This is the linear decrease schedule. We maintain a constant CMR dropout value for the first few epochs. We start decreasing the CMR value linearly to zero as soon as we start pruning $W_{s}$. We linearly decrease CMR to zero. The number of steps in which the CMR decreases to zero is equal to number of steps taken to prune $W_{s}$ to the required sparsity values. Finally, we train the network with no CMR (zero CMR) for a few more epochs.
    \item \textbf{expDec}: This is the exponential decrease schedule. We maintain a constant CMR dropout value for the first few epochs. We start decreasing the CMR value proportional to the density of the $W_{s}$ matrix as soon as we start pruning $W_{s}$. Once the $W_{s}$ matrix is pruned to the required sparsity value, CMR dropout converges to zero. Finally, we train the network with no CMR (zero CMR) for a few more epochs.
\end{itemize}

\begin{table}[t]
\centering
\caption{Ablation study comparing various CMR dropout schedules (Section \ref{sec:ablcmr}). Schedules that adapt to changing sparsity levels in $W_{s}$ achieve better results than other schedules. Thus, \textbf{linDec} and \textbf{expDec} achieve better Test Perplexity than \textbf{constant}. \textbf{constant} schedule maintains a large dropout value even after most of the weights in $W_{s}$ are pruned away, inhibiting learning.}
\label{tab:cmrsched}
\vspace{5pt}
\begin{tabular}{@{}clll@{}}
\toprule
\multicolumn{1}{l}{} & Compression & \begin{tabular}[c]{@{}l@{}}CMR\\Schedule\end{tabular} & \begin{tabular}[c]{@{}l@{}}Test\\Ppl.\end{tabular} \\ \midrule
\multicolumn{1}{l}{Baseline} & 1x & - & 82.1 \\
\multirow{3}{*}{Doped KP} & 20x & \textbf{constant} & 88.5 \\
 & 20x & \textbf{expDec} & 83.6 \\
 & 20x & \textbf{linDec} & 82.9 \\ 
\bottomrule
\end{tabular}
\end{table}

Results in Table \ref{tab:cmrsched} indicate that \textbf{linDec} achieves the best accuracy when compared to other schedules. \textbf{constant} achieves the highest test perplexity value while \textbf{expDec} achieves comparable accuracy to \textbf{linDec}. \textbf{constant} leads to least accuracy as having a CMR dropout after pruning away most of the weights in $W_{s}$ inhibits learning. Thus, schedules that adapt to the sparsity levels of $W_{s}$ generally fare better. This paper uses the \textbf{linDec} schedule for CMR dropout. 

\section{Results}
\label{sec:results}
\subsection{Datasets and Benchmarking Methodology}

We evaluate the compression technique on networks trained on two different datasets. We train the language models on the Penn Treebank Corpus \cite{ptbdataset}. The dataset consists of 929k training words, 73k validation words and 82k test words with a total vocabulary size of 10000. The machine translation network is trained on the  English-Vietnamese Translation dataset. The dataset consists of 133k training sentence examples, 1553 sentences in the validation set and 1268 sentences in the test set. The networks were trained using Tensorflow 1.14 platform using 2 Nvidia RTX 2080 GPUs.

We compare networks compressed using DKP with multiple alternatives: 
 \begin{itemize}
    \item \textit{Pruning}: We use the magnitude pruning framework provided by \cite{suyog}. While there are other possible ways to prune, recent work \cite{bestprune} has suggested that magnitude pruning provides state-of-the-art or comparable performance when compared to other pruning techniques \cite{varprune,l0prune}. 
    \item \textit{Low-rank Matrix Factorization (LMF)}: LMF ~\cite{DBLP:journals/corr/KuchaievG17} expresses a matrix $A \in \mathbb{R}^{m\times n}$ as a product of two matrices $U \in \mathbb{R}^{m \times d}$ and $V \in \mathbb{R}^{d \times n}$, where $d$ controls the compression factor. 
   \item \textit{Small Baseline}: Additionally, we train a smaller baseline with the number of parameters equal to that of the compressed baseline. The smaller baseline helps us evaluate if the network was over-parameterized.  
   \item \textit{Previous state-of-the-art:} Additionally, we show results comparing our method with previous state-of-the-art compression results on the same benchmark.
\end{itemize}

In order to understand the generality of the compression technique, we evaluate and compress 4 different benchmarks across two applications - Language Modeling and Language Translation. They are:
\begin{itemize}
    \item Medium LM in ~\cite{ptblm}: The LM consists of 2 LSTM layers with hidden size of 650. The baseline network was trained using a learning rate of 1.0 for 39 epochs with a learning rate decay of 0.8. For regularization we use max grad norm value of 5 and a dropout of 0.5 for all the layers.
    \item Large LM in ~\cite{ptblm}: The LM consists of 2 LSTM layers with hidden size of 1500. The baseline network was trained using a learning rate of 1.0 for 55 epochs with a learning rate decay of 0.85. For regularization we use max grad norm value of 10 and a dropout of 0.65 for all the layers. 
    \item RHN LM in ~\cite{rhn}: We train the RHN network with a depth of 10 layers and hidden size of 830. The input and output weight embedding are tied together. 
    \item En-Vi translation network in ~\cite{nmt_vien}: The model uses 2-layer LSTMs of size 512 units with bidirectional encoder and unidirectional decoder, embedding of dimension 512 and an attention layer. The network is trained using a learning rate of 1.0, using a dropout value of 0.2 and gradient norm value of 5.0.
\end{itemize}

For all of the above networks we compress the LSTM layers in the network, unless stated otherwise. We compare the accuracy of the compressed networks and identify the compression method that achieves the lowest preplexity (highest accuracy). The hyperparameters of the compressed network can be found in Appendix \ref{app:hp}. We also measure the number of operations required to execute the compressed network and the wall-clock inference run-time of the compressed network on an embedded device. 

\begin{table}[tb]
\caption{
Medium LM LSTM results demonstrate that doping is beneficial when applied to structured compression techniques, such as KP, LMF and HMD shown here. Doped structured matrix compression technique outperforms traditional structured matrix compression (w/o doping). Amongst the different structured matrices, KP outperforms LMF and HMD, due to the higher rank of KP structure. Additionally, the results also indicate that doped structured matrix trained without CMR do no achieve good accuracy due to CMA, thus indicating that CMA and CMR is a phenomenon unique to other structured matrices also.
}
\vspace{5pt}
\label{tab:dopedlmf}
\centering
\small
\begin{tabular}{@{}llll@{}}
\toprule
\multicolumn{1}{c}{Method} & \multicolumn{1}{c}{Compression} & \multicolumn{2}{c}{Test Perplexity} \\ 
 &  & Conventional & Doped \\
\midrule
Baseline & 1x & 82.1 & - \\ \cmidrule(r){1-4}
KP (no CMR) & 20x & 89.1 & 93.3 \\ 
KP w/ CMR & 20x & 89.1 & 82.9 \\ \cmidrule(r){1-4}
LMF (no CMR) & 20x & 103.4 & 107.1 \\ 
LMF w/ CMR & 20x & 103.4 & 89.2 \\ \cmidrule(r){1-4}
HMD (no CMR) & 20x & 98.7 & 104.8 \\
HMD w/ CMR & 20x & 98.7 & 87.4 \\ \bottomrule
\end{tabular}
\end{table}

\subsection{Impact of doping structured matrices}
\label{sec:dopedlmf}
Doping is applicable to a variety of structured matrices. 
We applied doping to low-rank matrix factorization (doped LMF) technique, HMD \cite{urmtha01-hmd} (doped HMD) and kronecker product (doped KP), using the CMR training method.
Table \ref{tab:dopedlmf} summarizes the results for standard and doped structured matrices, trained with and without CMR.
At the same compression ratios, 
doped structured matrix compression outperforms standard structured matrix compression by a margin of 14\% or more. This shows that doping, CMA and CMR are generally applicable to structured matrices also, opening opportunities for development and exploration of a broad range of doped structured compression methods. 

The results show that the structured matrix used has a significant influence on the final accuracy. Overall, we found a strong correlation between the rank of the structured matrix and the test perplexity results of the doped Structured Matrices. Doping KP matrices leads to superior accuracy than doping LMF and HMD structures. The KP structured matrix has an order of magnitude higher rank than that of LMF and HMD \cite{urmkron2} for same number of parameters. Therefore,the KP matrix is far more expressive than an LMF or HMD equivalent, resulting in superior performance. This makes KP a better structure to combine with doping. Similarly, HMD has double the rank of the matrix than one decomposed using LMF for the same number of parameters, thus doped HMD has a better perplexity than doped LMF. 

\subsection{Compression using Doped Kronecker Product matrices}
Section \ref{sec:dopedlmf} showed that doped KP compression outperforms other doped structured matrix based compression techniques by a large margin. In this section we will compare this compression technique with other strong alternatives on a wide variety of benchmarks.

\begin{figure}[tb]
\centering
\hspace{-0.5em}\includegraphics[width=\columnwidth]{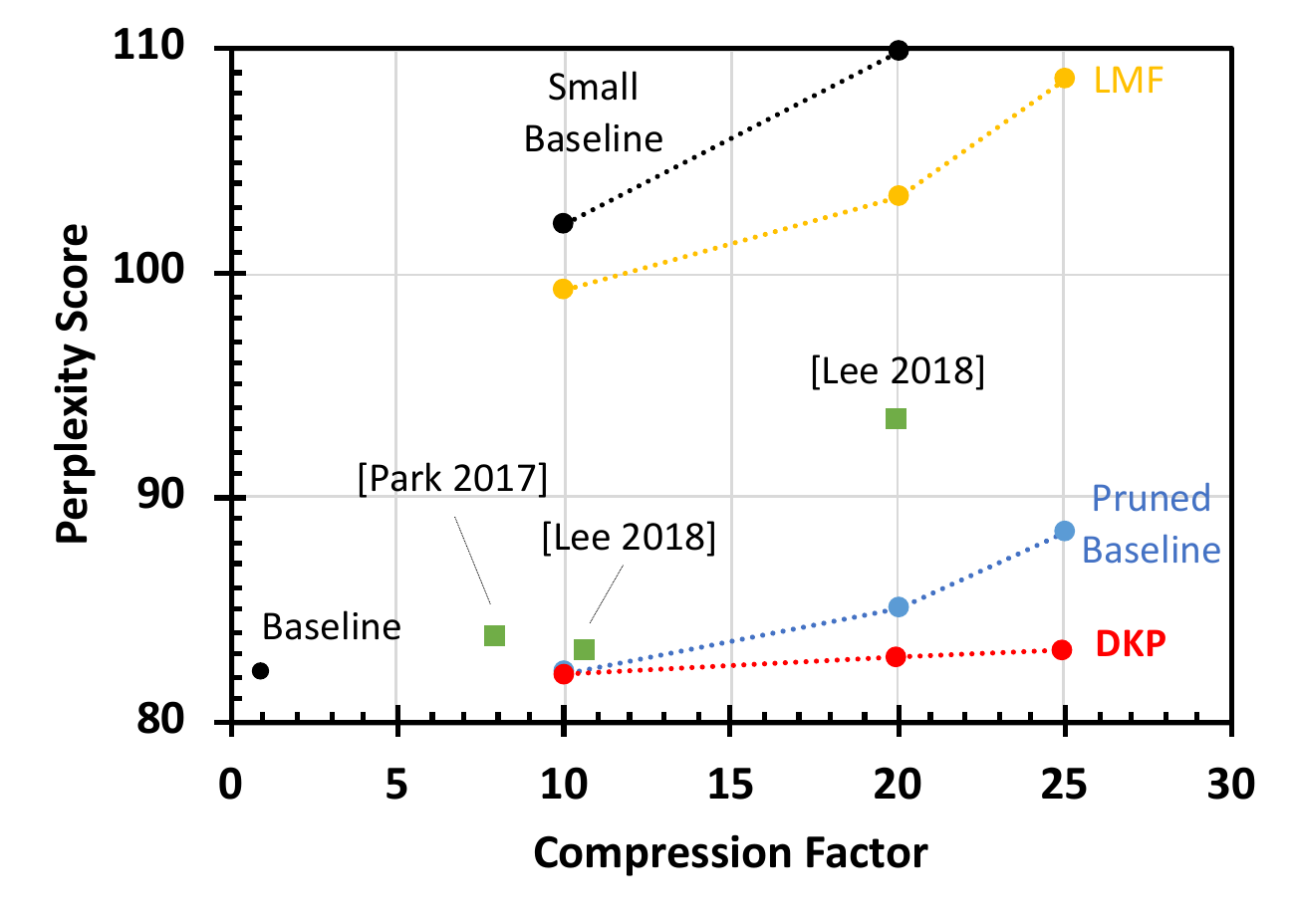}
\vskip -0.2in
\caption{
Medium LM at various compression factors. 
The baseline model defines $1\times$ compression. 
Doped KP is compared with pruning, LMF (\cite{DBLP:journals/corr/KuchaievG17}), a smaller baseline (SB), and previous published results on this benchmark \cite{2017_c1,2018_c2,2019_c4}.}
\label{fig:dkpScatter}
\end{figure}

\begin{table}[t]
\centering
\caption{Large LM results show that doped KP has the best perplexity at high compression factors, compared to previous work.} 
\vspace{5pt}
\label{tab:ptblarge}
\begin{tabular}{l c c}
\toprule
Method  & Compression & Test Perplexity \\ \midrule
Baseline Model & $1\times$ & 78.3 \\
\cite{iss} & $9.83\times$ & 78.6 \\
\cite{DBLP:journals/corr/abs-1906-06847} & $10.71\times$ & 78.1 \\
\cite{suyog} & $20\times$ &  83.4\\
Doped KP (This work) & $20\times$ & 78.5 \\
\bottomrule
\end{tabular}
\end{table}

\begin{figure}[t]
\centering
\hspace{-0.5em}\includegraphics[width=\columnwidth]{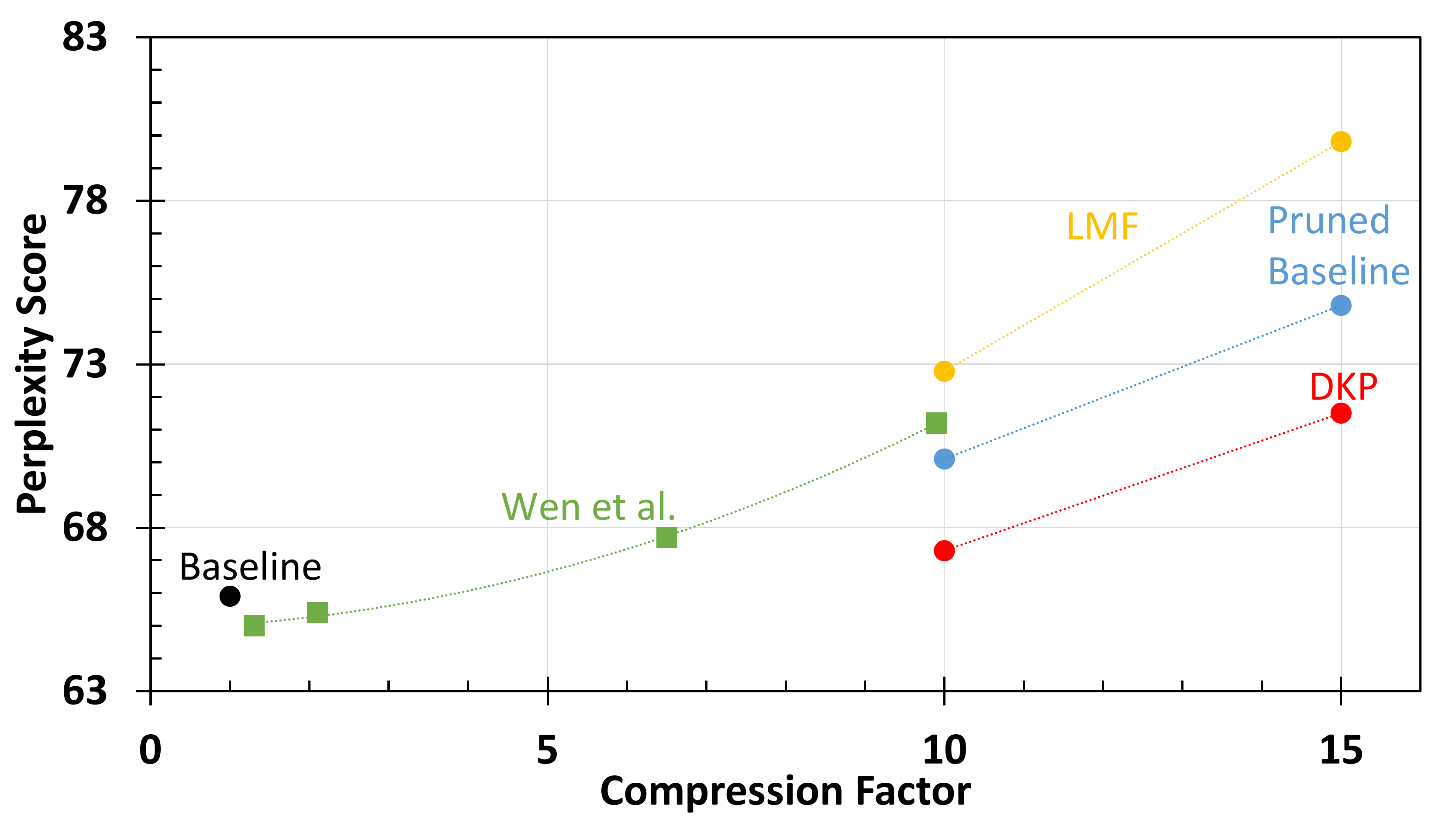}
\caption{
RHN LM at various compression factors. 
The baseline model defines $1\times$ compression. 
Doped KP is compared with pruning, LMF (\cite{DBLP:journals/corr/KuchaievG17}), and previous published results on this benchmark \cite{iss}.}
\label{fig:dkprhn}
\end{figure}

\begin{table}[t]
\centering
\caption{Compressing a more regularized Large LM regularized using doped KP. We use weight-tying regularization that serves dual purpose of compressed word embedding representation and more regularized network. The results indicate that doped KP can compress a more regularized version of Large LM.} 
\vspace{5pt}
\label{tab:ptblargereg}
\begin{tabular}{@{}llll@{}}
\toprule
Network & \multicolumn{2}{c}{Compression Factor} & Perplexity \\ 
 & \begin{tabular}[l]{@{}l@{}}Word\\Embeddings\end{tabular} & \begin{tabular}[l]{@{}l@{}}LSTM\\Layers\end{tabular} &  \\ \midrule
Large LM & 1x & 1x & 78.3 \\
+Weight Tying & 2x & 1x & 73.8 \\
+Doped KP & 2x & 15x & 76.2 \\ \bottomrule
\end{tabular}
\end{table}

\begin{table}[tb]
\centering
\caption{Results of Compression of a GNMT network}
\vspace{5pt}
\label{tab:gnmt}
\begin{tabular}{@{}lll@{}}
\toprule
\multicolumn{1}{c}{Method} & \multicolumn{1}{c}{Compression} & \multicolumn{1}{c}{\begin{tabular}[c]{@{}c@{}}BLEU\\ Score\end{tabular}} \\ \midrule
Baseline Model & 1x & 25.5 \\
Pruned Baseline & 15x & 24.2 \\
LMF & 15x & 22.8 \\
Small Baseline & 15x & 23.1 \\
Doped KP & 15x & 24.9 \\ \bottomrule
\end{tabular}
\end{table}

\subsubsection{Medium and Large LM in \cite{ptblm}}
\label{res:mediumlm}

Figure~\ref{fig:dkpScatter} shows the Medium LM doped KP results at
multiple compression factors, compared with pruned baselines (\cite{suyog}), LMF (\cite{DBLP:journals/corr/KuchaievG17}), and a small baseline with reduced layer sizes. 
We also include recently published results for the same LMs. 
As shown, doped KP outperforms all traditional compression techniques known at $20\times$ compression factors and beyond. 
These models have $W_{s}$ sparsity of $95\%$ or higher. 
Doped KP achieves 6\% better accuracy than pruning and 23.3\% better accuracy than LMF for $25\times$ compression factor. 
Additionally, doped KP outperforms all previously published results, with $2.4\times$ higher compression at the same perplexity as previous best result~\cite{2018_c2}.

Table~\ref{tab:ptblarge} shows the large LM model compressed using doped KP at 25$\times$ compression. 
Doped KP consistently outperforms both standard techniques and previous work. 
We improve the state-of-the art, almost doubling ($1.8\times$) the compression at comparable accuracy \cite{DBLP:journals/corr/abs-1906-06847}.

\begin{table*}[t]
\centering
\caption{Overview of the results highlighting the compressing abilities of doped KP method across a variety of benchmarks. Speed-up over baseline is measured by running the network on a Raspberry Pi 4 board using Eigen C++ Library.}
\vspace{5pt}
\label{tab:macresults}
\begin{threeparttable}
\centering
\begin{tabular}{@{}llcccc@{}}
\toprule
Benchmark & \begin{tabular}[c]{@{}l@{}}Compression\\ Technique\end{tabular} & \begin{tabular}[c]{@{}c@{}}Perplexity/ BLEU \\ score\end{tabular} & \begin{tabular}[c]{@{}c@{}}Compression\\ Factor\end{tabular} & \begin{tabular}[c]{@{}c@{}}MAC\\ Reduction\end{tabular} & \begin{tabular}[c]{@{}c@{}}Speed-up on \\ commodity hardware\end{tabular} \\ \midrule \cmidrule(l){1-6} 
\multirow{3}{*}{Medium LM} & Baseline & 82.1 & 1x & 1x & 1x \\
 & \cite{2019_c4} & 83.1 & 10x & Not Reported & NA\tnote{a} \\
 & Ours (Doped KP) & 83.2 & 25x & 7.41x & 4.06x \\ \cmidrule(l){1-6} 
\multirow{3}{*}{Large LM} & Baseline & 78.3 & 1x & 1x & 1x \\
 & \cite{DBLP:journals/corr/abs-1906-06847} & 78.1 & 10.71x & 7.48x & 14.7x \\
 & Ours (Doped KP) & 78.5 & 20x & 5.64x & 5.49x \\ \cmidrule(l){1-6} 
\multirow{3}{*}{RHN} & Baseline & 65.9 & 1x & 1x & 1x \\
 & \cite{iss} & 71.2 & 10x & 5.3x & 10.7x \\
 & Ours (Doped KP) & 71.1 & 15x & 5.79x & 5.34x \\ \cmidrule(l){1-6} 
\multirow{3}{*}{GNMT} & Baseline & 25.5 & 1x & 1x & 1x \\
 & \cite{suyog} & 24.9 & 10x & 10x & 3.3x \\
 & Ours (Doped KP) & 24.9 & 15x & 6.12x & 2.55x \\ 
 \bottomrule
\end{tabular}
        \begin{tablenotes}
            \item[a] Previous state-of-the-art~(\cite{2019_c4}) cannot be implemented on commodity hardware.
        \end{tablenotes}
\end{threeparttable}
\end{table*}

\textit{Impact of additional regularization:} We wanted to understand whether adding more regularization to the previous models negatively impacted the conclusions in the previous section. In order to do that, we regularized the Large LM using weight tying regularization \cite{weighttying}. Weight Tying (WT) ties the input and output word embeddings of a LM resulting in a smaller network with better generalization. As shown in Table \ref{tab:ptblargereg}, WT compresses the word embeddings by $2\times$ while simultaneously improves the perplexity of the Large LM by 4.5 points when compared to the baseline. Doped KP can further compress the network by $15\times$ with minor impact on perplexity score. 

\subsubsection{Recurrent Highway Networks (RHN)}
\label{res:mediumrhn}
We also tested Doped KP compression of a more recent and more heavily regularized LM by \citet{rhn}. Figure \ref{fig:dkprhn} shows the results of these experiments. As shown in the figure, doped KP can compress the RHN network by $10\times$ with only minor degradation in perplexity score. Additionally, doped KP achieves 1.5$\times$ more compression than previous state-of-the-art compression technology while achieving similar perplexity.

\subsubsection{Compressing a Language Translation Network}
\label{res:gnmt}
To understand whether doped KP can be used to compress applications beyond LMs, we compress an English-Vietnamese translation network from~\citet{nmt_vien}. Results in Table \ref{tab:gnmt} show that doped KP can achieve better accuracy than other traditional compression techniques. 

\subsubsection{Compressing an Intent Detection Network}
Using doping, we are able to compress the Intent Detection network \cite{DBLP:journals/corr/LiuL16d} trained on the ATIS dataset by 10 $\times $, with 0.7\% loss in baseline accuracy (97.65\%). Pruning the baseline network to a compression factor of 10$\times $ leads to 1.3\% loss in baseline accuracy, while LMF leads to 4.2\% loss in baseline accuracy.

\subsubsection{Impact on Op count and Inference run-time}
\label{res:inf}
Table \ref{tab:macresults} shows the multiply-accumulate operation (MAC) reduction achieved by the networks discussed in this paper for various compression factors. Doped KP achieved near iso-accuracy at $25\times$ compression for the Medium LM, $20\times$ for the Large LM, $15\times$ for GNMT and $10\times$ for RHN. These compression factors correspond to $7.41\times$, $5.64\times$, $6.12\times$  and $5.79\times$ reduction in MAC operations required to execute the LSTM layers of the network. 

To measure the wall-clock inference run-time, we implement the compressed network on a Raspberry Pi 4 board \cite{rasp}. The borad has Arm Cortex A-72 processor and the networks were implemented using Eigen C++ library \cite{eigen}. As shown in Table \ref{tab:macresults}, doped KP compressed networks not only achieve MAC reduction, but also achieves $2.5\times - 5.5\times$ inference speed-up. 

\section{Doped Quantized Network}

In general, the doping technique introduced in our paper should apply to any models that use fully connected layers and convolutions. Therefore, in addition to LSTMs, it should also be applicable to MLPs, transformers and CNNs. To demonstrate this, we ran further experiments to apply doping to a quantized AlexNet CNN \cite{NIPS2012_c399862d} trained on the ImageNet dataset \cite{5206848}. To do this, we first aggressively quantized the weights in an AlexNet model to 2-bits, which leads to a top-1 accuracy of 50.1\%. Next, we doped the network with 1\% of FP32 values. We found that this tiny level of doping increases the accuracy by 3.1\% to 53.1\%. Although this is a preliminary result, we believe it indicates that doping (i.e., the addition of sparse additive matrices) can be useful for CNNs, as well as the LSTMs studied in our paper. It further shows that doping can generalize to structures beyond those discussed in the paper.

\section{Limitation and Future Work}

In common with many other structured matrix approaches, Doped KP currently introduces a 2-3x training slowdown. This is due to (a) the additional memory required for the sparse matrix (which is initially dense before pruning), and (b) the lack of efficient KP GPU kernels (for both forward- and back-propagation). This training time limitation has so far prevented us from applying doping to big models and big datasets. We believe that (a) could potentially be addressed by the use of second order approximation methods \cite{NEURIPS2019_3a01fc08,NEURIPS2020_d77c7035}. These methods can help identify locations in the structured matrix that are stuck in sub-optimal minima and might benefit from a non-zero value in an equivalent location in the sparse matrix. This might allow us to learn the sparse matrix directly, without the need to prune down from a dense matrix. Additionally, (b) can be avoided by developing specialized GPU kernels for structured matrices. Recently, \cite{DONG2020102701} showed how to develop such kernels for block circular matrices, another popular structured matrix. We leave the training time challenge for future work.

\section{Conclusion}

This paper introduces doping, a technique to improve accuracy of networks compressed using structured matrices like the Kronecker product (KP). 
Doping introduces an additional additive matrix to the the KP matrices, which we force to be very sparse during training. To train high accuracy networks with large compression factors, doped structured matrices need to over-come co-matrix adaptation (CMA) using the co-matrix regularization (CMR). Doped structured matrices trained using CMR and CMA can train to higher accuracy at larger compression factors than using the structured matrix alone, while achieving better compression factors than previous state-of-the-art compression technique. Specifically, doped KP lead to $10\times - 25\times$ with minor loss in accuracy while improving the compression factor of previous state-of-the-art by $1.3\times - 2.4\times$. These results were collected by evaluating 4 different NLP application in the domain of language modeling and language translation. Additionally, doped KP compressed network can be deployed on commodity hardware achieving inference speed-up of $2.5-5.5\times$ over baseline.
\clearpage
\bibliographystyle{mlsys2020}
\bibliography{main}

\clearpage
\appendix
\section{Impact of the dimensions of B and C matrices in $W_{k}$ }
\label{sec:kpchoice}

\begin{table}[tbh]
\caption{Impact of the choice of different KP compression methodology used for $W_{k}$ matrix. In this paper we follow the methodology in \cite{urmkron2}. The results of this ablation study compare this choice with various alternatives discussed in \cite{urmkron2}. For overall compression of $10\times$, we can either start from a $338\times$ KP compressed method and add a $W_{s}$ matrix with $9\%$ non-zero values ( $9\%$ doping) or start from one of the other alternative points and vary the amount of doping. The former follows the methodology used in this paper, while the later describes alternative methodologies that could have been followed. These results validate our initial choice of KP compression for $W_{k}$ and show that the methodology followed in this paper is optimal.}
\label{tab:ablkpmeth}
\begin{tabular}{lllll}
\toprule
\multicolumn{1}{c}{Methodology} & \multicolumn{1}{c}{\begin{tabular}[c]{@{}c@{}}KP \\ Compre-\\ ssion\end{tabular}} & \multicolumn{1}{c}{\begin{tabular}[c]{@{}c@{}}Doping\\ \%\end{tabular}} & \multicolumn{1}{c}{\begin{tabular}[c]{@{}c@{}}Overall\\ Compre-\\ ssion\end{tabular}} & \multicolumn{1}{c}{\begin{tabular}[c]{@{}c@{}}Test \\ Ppl.\end{tabular}} \\ \midrule
Alternative 1 & 10$\times$ & 0\% & 10$\times$ & 97.4 \\
Alternative 2 & 20$\times$ & 4.5\% & 10$\times$ & 91.7 \\
Alternative 3 & 40$\times$ & 7\% & 10$\times$ & 86.8 \\
\midrule
\begin{tabular}[c]{@{}l@{}}This\\ Paper\end{tabular} & 338$\times$ & 9\% & 10$\times$ & 82.1 \\ \bottomrule
\end{tabular}
\end{table}

Section \ref{sec:kpparam} discussed the various choices for the dimension of $B$ and $C$ matrices to express $W_{k}$. For example, if $\boldsymbol{W_{k}}$ is of size 100$\times$100, B and C can be of size $2\times50 and 50\times 2$ each or $10\times10 and 10\times 10$ each. In this paper, we use the methodology proposed in \cite{urmkron2} to identify the configuration of $B$ and $C$ matrix that achieves maximum compression while still preserving the rank of the matrix after compression. Their results indicate that this methodology leads to better accuracy than others. We run a study to validate whether their assumption is applicable to Doped KP networks and thus to validate our initial choice for setting up the compression problem in section \ref{sec:kpparam}. Table \ref{tab:ablkpmeth} shows the results for this ablation study. The various rows in the table indicate the various methodologies for KP compression and ways to achieve 10$\times$ compression using them. For example the last row shows that using the methodology in this paper, we can start with a $W_{k}$ matrix that is compressed by $338\times$ and add a $W_{s}$ matrix with $9\%$ non-zero parameters ($9\%$ doping) to achieve $10\times$ overall compression. The second last row starts with one of the alternative KP compression methods that compresses the $W_{k}$ matrix by $40\times$ and dopes $7\%$ parameters in the $W_{s}$ matrix to achieve an overall compression of $10\times$. The results validate the choices made in the paper. For iso-compression, doping on top of a matrix compressed using KP compression methodology in \cite{urmkron2} achieves least perplexity score when compared to doping on top of a matrix compressed using alternative KP compression methodologies. 

\section{Hyper-parameters}
\label{app:hp}
\label{sec:appendixHP}
\begin{table}[htb]
\caption{Large LM Hyper-parameters}
\label{tab:lmhype}
\begin{tabular}{|l|l|c|c|}
\hline
\multicolumn{2}{|l|}{\begin{tabular}[c]{@{}l@{}}Large LM \\ Hyper-parameters\end{tabular}} & Baseline & \begin{tabular}[c]{@{}c@{}}20x compressed\\ network\end{tabular} \\ \hline
\multicolumn{2}{|l|}{size(W)} & \multicolumn{2}{c|}{6000x3000} \\ \hline
\multicolumn{2}{|l|}{size(Wk)} & NA & \begin{tabular}[c]{@{}c@{}}52x65\\ 50x20\end{tabular} \\ \hline
\multicolumn{2}{|l|}{sparsity(Ws)} & NA & 96.60\% \\ \hline
\multicolumn{2}{|l|}{Initial LR} & 1 & 0.3 \\ \hline
\multicolumn{2}{|l|}{LR decay} & 0.85 & 0.96 \\ \hline
\multicolumn{2}{|l|}{\#Epochs} & 55 & 100 \\ \hline
\multicolumn{2}{|l|}{\begin{tabular}[c]{@{}l@{}}LR Decay \\ Start Epoch\end{tabular}} & 10 & 15 \\ \hline
\multicolumn{2}{|l|}{CMR} & NA & 0.7 \\ \hline
\multicolumn{2}{|l|}{L2 Regularization} & 0.0001 & 0.0001 \\ \hline
\multicolumn{2}{|l|}{Max Grad Norm} & 5 & 5 \\ \hline
\multicolumn{2}{|l|}{Dropout} & 0.65 & 0.65 \\ \hline
\multicolumn{1}{|c|}{\multirow{2}{*}{\begin{tabular}[c]{@{}c@{}}Sparsity \\ Schedule\end{tabular}}} & \begin{tabular}[c]{@{}l@{}}Epoch \#\\ during start\\ of pruning\end{tabular} & NA & 20 \\ \cline{2-4} 
\multicolumn{1}{|c|}{} & \begin{tabular}[c]{@{}l@{}}Epoch \#\\ during end\\ of pruning\end{tabular} & NA & 90 \\ \hline
\end{tabular}
\end{table}

\begin{table}[!htb]
\caption{GNMT Hyper-parameters}
\label{tab:gnmthyp}
\begin{tabular}{|l|l|c|c|}
\hline
\multicolumn{2}{|l|}{} & Baseline & Compressed \\ \hline
\multicolumn{2}{|l|}{\begin{tabular}[c]{@{}l@{}}Attention\\ Type\end{tabular}} & \multicolumn{2}{c|}{scaled luong} \\ \hline
\multicolumn{2}{|l|}{dropout} & \multicolumn{1}{l|}{0.2} & \multicolumn{1}{l|}{0.2} \\ \hline
\multicolumn{2}{|l|}{encoder type} & \multicolumn{2}{c|}{bidirectional} \\ \hline
\multicolumn{2}{|l|}{learning rate} & 1.0 & 1 \\ \hline
\multicolumn{2}{|l|}{max grad norm} & 5 & 5 \\ \hline
\multicolumn{2}{|l|}{size(W)} & \multicolumn{2}{c|}{2048x1024} \\ \hline
\multicolumn{2}{|l|}{size(Wk)} & NA & \begin{tabular}[c]{@{}c@{}}32x8\\ 64x128\end{tabular} \\ \hline
\multicolumn{2}{|l|}{sparsity(Ws)} & NA &  \\ \hline
\multicolumn{2}{|l|}{src max len} & 50 & 50 \\ \hline
\multicolumn{2}{|l|}{beam width} & 10 & 10 \\ \hline
\multicolumn{2}{|l|}{\#TrainSteps} & 12000 & 20000 \\ \hline
\multicolumn{2}{|l|}{LR Schedule} & \multicolumn{2}{c|}{luang234} \\ \hline
\multirow{2}{*}{\begin{tabular}[c]{@{}l@{}}Sparsity\\ Schedule\end{tabular}} & \begin{tabular}[c]{@{}l@{}}Epoch \#\\ during start\\ of pruning\end{tabular} & 2000 & 130000 \\ \cline{2-4} 
 & \begin{tabular}[c]{@{}l@{}}Epoch \#\\ during end\\ of pruning\end{tabular} & 2000 & 130000 \\ \hline
\end{tabular}
\end{table}

\begin{table*}[htb]
\centering
\caption{Medium LM hyperparameters}
\label{tab:medlmhyp}
\begin{tabular}{|l|l|c|c|c|c|}
\hline
\multicolumn{2}{|l|}{\begin{tabular}[c]{@{}l@{}}Medium LM \\ Hyper-parameters\end{tabular}} & \multicolumn{1}{l|}{Baseline} & \multicolumn{1}{l|}{\begin{tabular}[c]{@{}l@{}}10x \\ compressed\\ network\end{tabular}} & \multicolumn{1}{l|}{\begin{tabular}[c]{@{}l@{}}20x \\ compressed\\ network\end{tabular}} & \multicolumn{1}{l|}{\begin{tabular}[c]{@{}l@{}}25x \\ compressed\\ network\end{tabular}} \\ \hline
\multicolumn{2}{|l|}{size(W)} & \multicolumn{4}{c|}{2600x1300} \\ \hline
\multicolumn{2}{|l|}{size(Wk)} & NA & \multicolumn{3}{c|}{52x65 \& 50x20} \\ \hline
\multicolumn{2}{|l|}{sparsity(Ws)} & NA & 91.10\% & 95.30\% & 96.30\% \\ \hline
\multicolumn{2}{|l|}{Initial LR} & 1 & 0.3 & 0.3 & 0.3 \\ \hline
\multicolumn{2}{|l|}{LR decay} & 0.8 & 0.96 & 0.96 & 0.96 \\ \hline
\multicolumn{2}{|l|}{\#Epochs} & 40 & 100 & 100 & 100 \\ \hline
\multicolumn{2}{|l|}{\begin{tabular}[c]{@{}l@{}}LR Decay \\ Start Epoch\end{tabular}} & 5 & 15 & 15 & 15 \\ \hline
\multicolumn{2}{|l|}{CMR} & NA & 0.7 & 0.7 & 0.7 \\ \hline
\multicolumn{2}{|l|}{L2 Regularization} & 0.0001 & 0.0001 & 0.0001 & 0.0001 \\ \hline
\multicolumn{2}{|l|}{Max Grad Norm} & 5 & 5 & 5 & 5 \\ \hline
\multicolumn{2}{|l|}{Dropout} & 0.5 & 0.5 & 0.5 & 0.5 \\ \hline
\multicolumn{1}{|c|}{\multirow{2}{*}{Sparsity Scedule}} & \begin{tabular}[c]{@{}l@{}}Epoch \# \\ during start\\ of pruning\end{tabular} & NA & 20 & 20 & 20 \\ \cline{2-6} 
\multicolumn{1}{|c|}{} & \begin{tabular}[c]{@{}l@{}}Epoch \# \\ during end\\ of pruning\end{tabular} & NA & 90 & 90 & 90 \\ \hline
\end{tabular}
\end{table*}

\end{document}